\documentclass[sigconf]{acmart}

\usepackage{booktabs} 
\usepackage{amssymb}

\begin{document}
\title{Occlusion Handling using Semantic Segmentation and Visibility-Based Rendering for Mixed Reality}

%

\author{Menandro Roxas}
\author{Tomoki Hori}
\author{Taiki Fukiage}
\author{Yasuhide Okamoto}
\author{Takeshi Oishi}
\affiliation{%
  \institution{The University of Tokyo}
  \city{Tokyo} 
  \state{Japan} 
}

\begin{abstract}
Real-time occlusion handling is a major problem in outdoor mixed reality system because it requires great computational cost mainly due to the complexity of the scene. Using only segmentation, it is difficult to accurately render a virtual object occluded by complex objects such as trees, bushes etc. In this paper, we propose a novel occlusion handling method for real-time, outdoor, and omni-directional mixed reality system using only the information from a monocular image sequence. We first present a semantic segmentation scheme for predicting the amount of visibility for different type of objects in the scene. We also simultaneously calculate a foreground probability map using depth estimation derived from optical flow. Finally, we combine the segmentation result and the probability map to render the computer generated object and the real scene using a visibility-based rendering method. Our results show great improvement in handling occlusions compared to existing blending based methods. 
\end{abstract}

%
%


\settopmatter{printfolios=true}

\maketitle

\section{Introduction}
In mixed reality, contradictory occlusion problem happens when a foreground real object is partially or completely covered by a background virtual object. To address this problem, the foreground real scene needs to be accurately segmented from the image frame. However, it is difficult to precisely extract the foreground in real time especially when the object is complex which are ubiquitous in outdoor scenes. Moreover, segmentation becomes more difficult when dealing with moving cameras because it has to be performed per frame and achieved in real-time. In this work, we focus on solving this occlusion problem specifically for an outdoor MR system on a moving vehicle using a monocular omnidirectional camera. 

Several methods have been proposed that handles foreground extraction in a monocular camera. Methods presented in \cite{grabcut} and \cite{lazysnap} can extract a very accurate foreground region from still images. However, extending the method to videos is inherently difficult due to computational cost of the segmentation technique used. In \cite{boundarymatting} and \cite{criminisi}, contours are cut by using alpha matting \cite{grabcut}, however an accurate prior foreground estimation is still required and the method fails even for small inaccuracy in the estimation especially for complex scenes such as tree branches and vegetation. 
\begin{figure}[t]
	\begin{center}
		\includegraphics[width=1.0\linewidth]{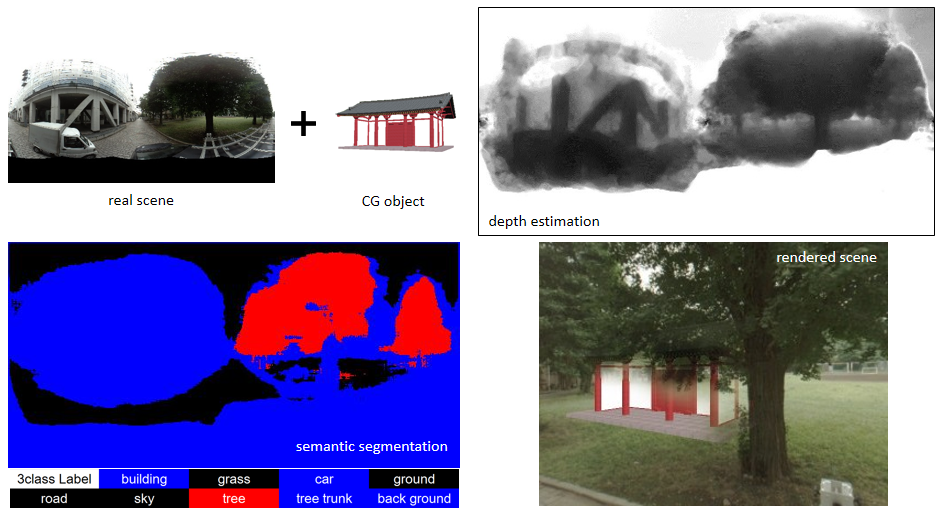}
	\end{center}
	\caption{Results of semantic segmentation, depth estimation and visibility-based rendering for occlusion handling in outdoor mixed reality.}
	\label{fig:occlusion}
\end{figure}

\begin{figure}[b]
	\begin{center}
		\includegraphics[width=1.0\linewidth]{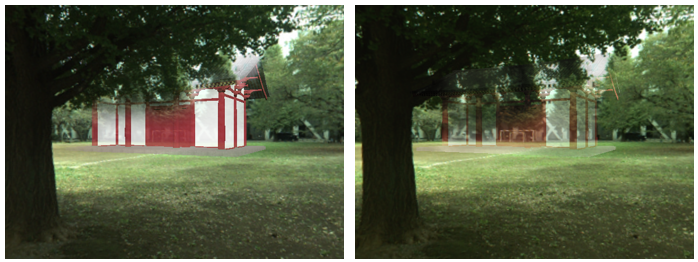}
	\end{center}
	\caption{Failure case of alpha blending (left: poor boundary handling) and transparency blending \cite{fukiage} (right: poor visibility) for handling occlusions.}
	\label{fig:error}
\end{figure}

\begin{figure*}
	\begin{center}
		\includegraphics[width=1.0\linewidth]{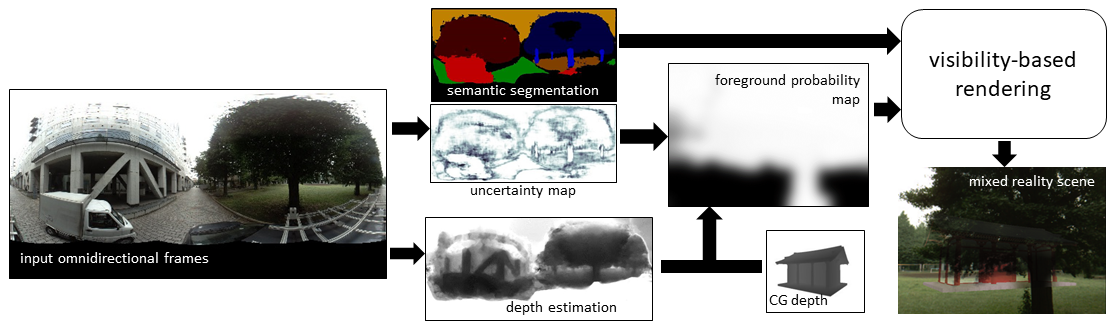}
	\end{center}
	\caption{Overview of the proposed method.}
	\label{fig:overview}
\end{figure*}

Background subtraction methods \cite{criminisi}\cite{backgroundcut} achieve real-time processing and can be applied for monocular cameras. In \cite{kakuta}\cite{vinh}, the background subtraction technique are modified and applied on an arbitrary outdoor environment taking advantage of multipile cues such as color, motion, and luminance change. However, these methods are constrained on a fixed camera and extension to moving camera applications is difficult.

Other methods use depth information \cite{depth1}\cite{depth2}\cite{depth3}\cite{depth4} to effectively reason on the foreground-background relationship of the virtual and real objects. By adding additional hardware such as multiple cameras for stereo vision, time-of-flight cameras, and range sensors, depth estimation is a straightforward foreground detection method that can be done in real-time. However, additional hardware is not always a desired solution in some applications.

Image-based methods in solving depth information for foreground extraction and segmentation in mixed reality field has been proposed before. In \cite{zollmann}, sparse 3D model data from a GIS were used to infer a dense depth map of the real scene. However, these prior 3D models are not easily available in most cases.

In this paper, we propose to use semantic segmentation for handling occlusions. We assign different attributes (i.e. amount of visibility, or transparency) depending on the class of an object. The reasoning is straightforward: for outdoor augmented reality, the sky and ground are background, and therefore should be hidden behind the CG object (or transparent). The rest could either be background or foreground. For objects that can be classified as both, we propose a real time foreground probability map estimation based on motion stereo. 

By combining the semantic segments and the probability map, we overlay the CG object onto the real scene by adapting a visibility-based rendering method first proposed in \cite{fukiage}. In \cite{fukiage}, a blending method, which uses visibility predictor based on human vision system, was used to predict the visibility (or transparency) level of the CG object. The method has an advantage over alpha blending methods (Figure \ref{fig:error}) because it does not require accurate estimation of complex boundaries. The method predicts the visibility of the CG object based on the color of the pixels and the foreground probability map inside a blending window. However, the blending method fails when the color of the foreground and background objects within the window are very similar, in which case the virtual object becomes too transparent (Figure \ref{fig:error}).

Instead of using a fixed visibility level for all objects, as in \cite{fukiage}, we use our proposed semantic classes to choose the amount of visibility for different type of objects. This allows us to control the appearance of the rendered object based on the type of the scene.

To summarize, this work has three main contributions (see Figure \ref{fig:overview}). First, we present a category scheme that uses semantics for assigning visibility values. We achieve this by first classifying the scene into specific categories using a convolutional neural network (CNN) based semantic segmentation method (SegNet \cite{segnet}). We then use our proposed scheme to group the segments into more usable categories to be used in visibility blending. Second, we present a real-time foreground probability map estimation method based on depth and optical flow for omnidirectional cameras.  Finally, we combine the semantic segmentation and the foreground probability map to create a visually pleasing augmented reality scene using visibility-based rendering. 

This paper is organized as follows. In Section \ref{sec:semantic}, we propose a category scheme for foreground prediction using semantic segmentation. In \ref{sec:foreground}, we present our foreground probability map estimation method. In Section \ref{sec:visibility} we introduce our visibility-based blending method which uses semantic classsification and the foreground probability map. In Section \ref{sec:results}, we show our results and comparison among our method, alpha blending, and transparency blending \cite{fukiage} methods. Finally, we conclude this work in Section \ref{sec:conclusion}.

\section{Semantic Segmentation for Occlusion Handling}
\label{sec:semantic}
\begin{figure}
	\begin{center}
		\includegraphics[width=1.0\linewidth]{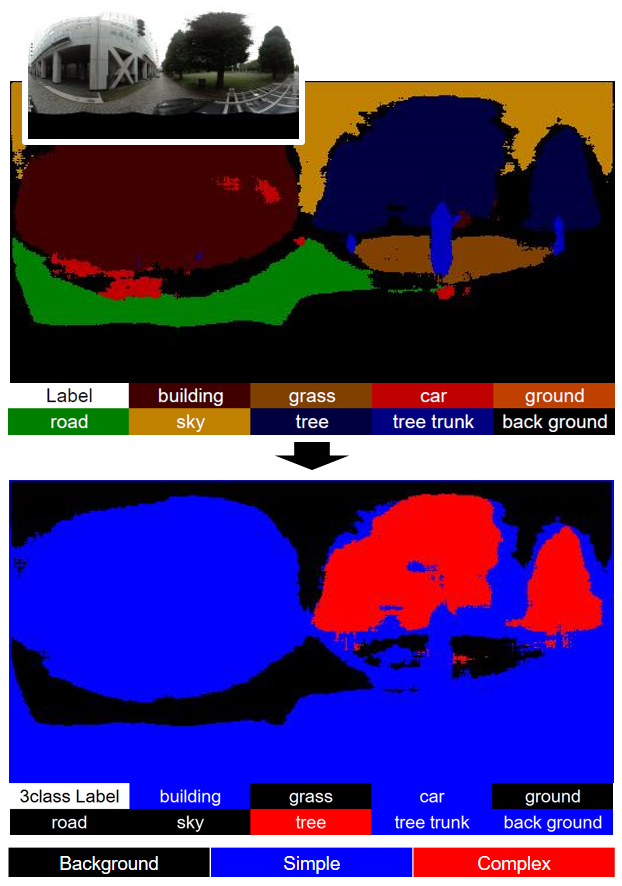}
	\end{center}
	\caption{First and Second stage of semantic segmentation. The first stage segments the scene into nine classes. The second stage groups the segmented classes into the three main categories.}
	\label{fig:semanticscheme}
\end{figure}
Given an image of the real scene, we need to categorize each objects in it as either foreground or background. Instead of directly using these two labels, we classify the object into three main categories: Background, Complex and Simple. Objects that belong to the Background category are those that are always in the background such as the ground or the sky, and are therefore transparent. On the other hand, objects that are classified as Complex or Simple can either be foreground or background, depending on the actual depth order of the CG object and the real scene. Complex and Simple objects are handled using the method we will describe in the following sections. 

In order to implement the above scheme, we first segment the scene into more specific categories. We use nine classes: Building, Grass, Car, Ground, Road, Sky, Tree, Tree trunk and Unknown. Note that the Tree category is mostly the leaves part of the tree (i.e. without the tree trunk). 

After this first stage of segmentation, we then group the resulting sections into our proposed main categories: Background (Grass, Ground Road, Sky, Unknown), Simple (Building, Car, Tree Trunk) and Complex (Tree) (see Figure \ref{fig:semanticscheme}). We use these three categories but additional classes can be added depending on the type of the scene where the mixed reality system is deployed.



This choice of implementation (two stage) is done to avoid misclassification which is possible when the class size is very small. For example, the Road and Grass section are visually different but they belong to the same Background category. Moreover, Grass, which is always in the background, is visually closer to a Tree, which can either be  in the foreground or background region. Therefore, we opt to allow the learning of the more refined classes instead of combining them into one semantic class. 

The result of the semantic segmentation are two outputs: the labeled segments and the uncertainty of prediction (see Figure \ref{fig:semanticsegmentation}). The uncertainty of prediction ($0 < g < 1$) is usually high along the object boundaries. We utilize this value to smoothly transit the visibility value between two different categories resulting in a more visually pleasing boundary handling (see Sec \ref{sec:visibility})

\section{Foreground Probability Map Estimation}
\label{sec:foreground}
For handling non-Background objects, we need to compare the actual depth of the scene and the rendered CG objects. To do this, we estimate a foreground probability map that indicates whether the real object is in the foreground or not. We solve this probability map by first estimating the depth of the scene for each frame, with respect to the center of the camera. 

\begin{figure}[t]
	\begin{center}
		\includegraphics[width=1.0\linewidth]{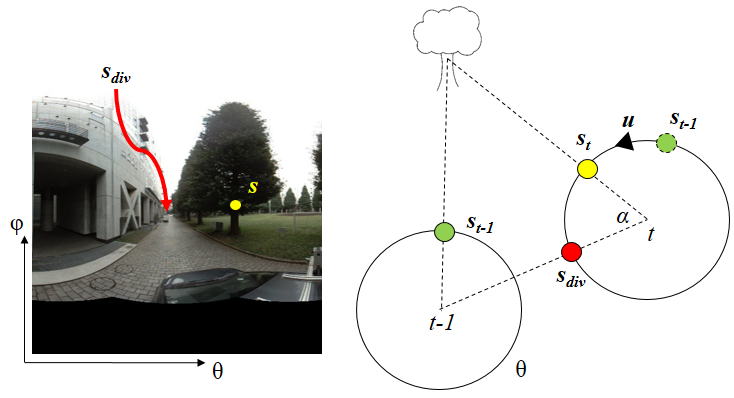}
	\end{center}
	\caption{Divergence point of the optical flow vectors and the relationship of the paralax angles of corresponding pixels $\mathbf{s_t, s_{t-1}}$ and the optical flow vector $\mathbf{u}$.}
	\label{fig:divergencepoint}
\end{figure}

Given an omnidirectional frame $I$ at time $t$, with pixel position corresponding to polar and azimuthal angles $\mathbf{s} = (\theta, \phi)$ and unity radial distance (focal length $f=1$), we first solve the optical flow vector $\mathbf{u} \in \mathbb{R}^2$ for every pixel $\mathbf{s}$. Assuming known camera positions $\mathbf{c}_{t}$ and $\mathbf{c}_{t-1}$, where $t-1$ is the position of the previous frame, and $\mathbf{c} \in \mathbb{R}^3$ in real-world coordinate system, we then solve the depth $d_{real}$ of $\mathbf{s}$ using triangulation:

\begin{equation}
\label{eq:triangulation}
d_{real}(\mathbf{s}) = |c - c_{t-1}| \frac{\sin{\alpha}}{\sin{\alpha}\cos{\alpha_{t-1}} - \cos{\alpha}\sin{\alpha_{t-1}}}
\end{equation}

where  $\alpha_t$ and $\alpha_{t-1}$ are the parallax angles calculated as the offset from corresponding pixels $\mathbf{s}_t$ and $\mathbf{s}_{t-1}$ to $\mathbf{s}_{div}$. $\mathbf{s}_{div} = ({\theta}_{div}, {\phi}_{div})$) is the direction of motion corresponding to the divergence point of the optical flow vectors. (See Figure \ref{fig:divergencepoint}). To solve the divergence point, we first estimate a rectangular region around the general direction of motion in the omnidirectional image. Since we know the position and orientation of the camera on the vehicle, we only need to extract the direction vector. Within this rectangular region in the image, we perform a convolution between the 2D optical flow vectors and a divergence kernel. The minimum value is then assigned as the divergence point.

\begin{figure}
	\begin{center}
		\includegraphics[width=1.0\linewidth]{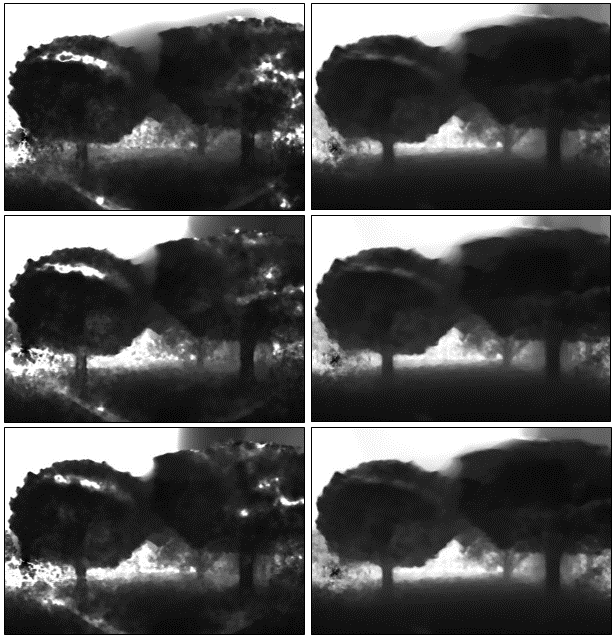}
	\end{center}
	\caption{Normalized depth (left column) and smoothed depth (right column).}
	\label{fig:depth}
\end{figure}

In order to handle the inaccuracy of the camera position estimation, we perform temporal smoothing of the depth of corresponding points in the image sequence along several consecutive frames. Since the optical flow is already given, we simply perform bilinear warping of the depth form one frame to another using the flow vectors as mapping. After warping the depth to the reference frame (current view), we simply averaged the depth values in the same pixel position. Figure \ref{fig:depth} shows the normalized depth and the temporally smoothed depth map of the real scene. 

Using $d_{real}$ and the corresponding depth of the virtual object $d_{cg}$, we calculate the foreground probability as:

\begin{equation}
\label{eq:foregroundprobability}
P_f = \frac{1}{1+e^{-(d_{cg}-d_{real})}}
\end{equation}

Equation \ref{eq:foregroundprobability} is a straightforward computation of the foreground probability map. The value is high if the depth of the real scene is smaller than that of the virtual object, which means that the real scene is closer to the camera. As the depth difference becomes smaller, however, the probability only decreases gradually so as not to suffer from inconsistency in depth estimation.

\section{Visibility-Based Rendering}
\label{sec:visibility}
We extend a visibility-based blending method \cite{fukiage2} to further utilize the semantic classification. This blending technique allows us to locally optimize the visibility of each region of the virtual object that can achieve arbitrarily trageted level. In \cite{fukiage2}, a visibility predictor based on human vision system is used in blending the virtual object. We extend this technique and use the semantic class and uncertainty of prediction in order to calculate the visibility value.

We first define the visibility of the virtual object $V_{cg}$ (as in \cite{fukiage} Eq. (7)) as:

\begin{equation}
\label{eq:visibility}
V_{cg} = (1-\omega)V_f + \omega V_b
\end{equation}  

From here, we deviate from \cite{fukiage} by setting:

\begin{equation}
\omega = \frac{1}{\sqrt{2 \pi \sigma}}\exp(-\frac{P_f^2}{2 \sigma})
\end{equation}

as the weighting based on the foreground probability map $P_f$. $V_{cg}$ is the weighted average between the foreground and background visibility values $V_f$ and $V_b$ and is averaged within a square window. $V_f$ is the visibility of the CG object when the real scene is a foreground, and $V_b$ is when the real scene is a background. We calculate these values based on the uncertainty $g$ and type of the segment classification:

	\begin{align}
	\label{eq:visibilities}
		V_f &= \frac{1}{2}V_{f1} + \frac{1}{2}\{(1-g)V_{f1} + gV_{f2}\}\\ \nonumber
		V_b &= \frac{1}{2}V_{b1} + \frac{1}{2}\{(1-g)V_{b1} + gV_{b2}\}
	\end{align}
	

where $V_{f1}$, $V_{f2}$, $V_{b1}$, and $V_{b2}$ are arbitrary values set by the user based on the desired appearance of the augmented scene depending on the class of the segment. $V_{f1}$ and $V_{b1}$ are the desired maximum visibility and $V_{f2}$ and $V_{b2}$ are the fallback minimum visibility. For the Background class, $V_f1$ and $V_b1$ are set to a high value such that the foreground probability map will be ignored. This is due to the fact that background object should not be visible.

For the Simple class, $V_{f1}$ is set to a very low value (almost zero), where as $V_{b1}$ is set to high value. In contrast, the Complex class has $V_{f1}$ also set to a high value, which should mean that when the Complex object is in the foreground the CG object will still be visible. This is not the case. In our observation, the Complex class tend to always appear in the foreground due to its texture complexity. Hence, when we solve $V_{cg}$ within the square window containing a Complex object, the CG object appears more transparent. We avoid this case by setting a high value for $V_{f1}$.

Equation \ref{eq:visibilities} allows gradual shifting from different visibility levels through the uncertainty value. This scheme is particularly effective along object boundaries. For example, if the uncertainty is very low (i.e. $g=0.01$) for a background object in the Simple category (i.e. Tree trunk), the visibility $V_f$ and $V_b$ is almost equavalent of the maximum visibility. In this case, if the foreground probibility map is high (i.e. $P_f = 0.95$), the total visibility of the CG object $V_{cg}$ approaches the maximum visibility for foreground $V_f$. On the other hand, if the uncertainty is high, (i.e. g = 0.85) which usually happens along boundaries, then $V_f$ and $V_b$ become weighted averages of the maximum visibility $V_{b1}$ and the fallback minimium visibility $V_{b2}$.

\section{Implementation}
\subsection{Test Environment}
Our method requires that the test environment and the dataset for semantic segmentation are similar. Due to this, we created a dataset of our test environment which consists of man-made objects such as buildings and cars and natural objects such as trees. Since we are using monocular depth estimation, we assume that no dynamic object are present (i.e., moving cars, walking people). Because we could not discriminate between static and dynamic objects, the computed depths are solved assuming that all objects in the scene are static. This will obviously result in wrong depth maps for dynamic objects (i.e. being farther or nearer due to motion parallax) but this is beyond the scope of this paper.

\subsection{Mixed Reality Bus System}
We tested our method on a Mixed Reality Bus (MR Bus) system. In this system, the real scene is captured by two omnidirectional cameras mounted externally on the vehicle. The two panoramic images are stitched together to form a single panoramic image. Each image is served on a computer which solves the depth estimation and another which performs the semantic segmentation. Another image, the composite depth map of the CG object is also solved given the current position of the bus. The results are then combined to create the foreground probability map.

\begin{figure}
	\begin{center}
		\includegraphics[width=1.0\linewidth]{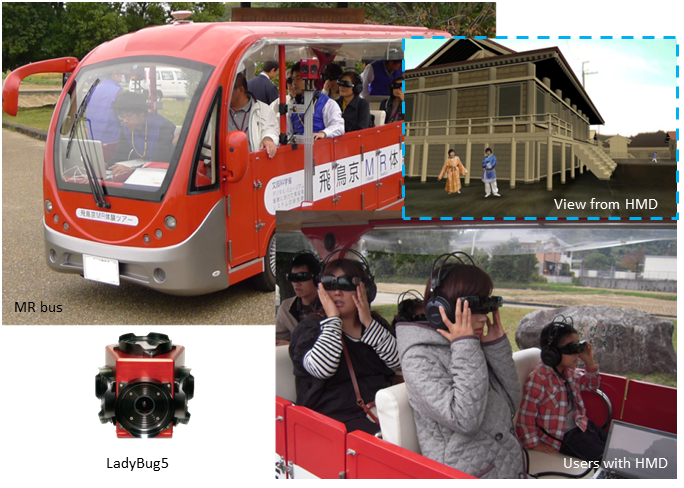}
	\end{center}
	\caption{Mixed Reality Bus Sytem. The scene is captured by an external camera, processed and rendered on the HMD.}
	\label{fig:mrsystem}
\end{figure}

The foreground probability map and the real scene are then served to the rendering system of the commercial-off-the-shelf head-mounted displays or HMDs. Using the HMD hardware (gyroscope and compass) for head pose estimation, it is straightforward to convert the panoramic images to the view of the HMDs (perspective) using image warping. Using the proposed blending method, the rendering system combines the perspective real scene and the CG objects. Figure \ref{fig:mrsystem} shows the actual MR bus used in our application. 

%
%
\subsection{SegNet}
For the semantic segmentation, we use the SegNet implementation provided by the authors \cite{segnetwebsite} compiled on a Intel Core i7-4930K, 32GB RAM and GTX980Ti GPU. We used the default Bayesian SegNet Basic model for our application.

For the dataset, we used omnidirectional images captured from a LadyBug\cite{ladybug} camera. Using the default resolution ($2048\times1024$), we manually created the labels using the open annotation tool LabelMe \cite{labelme}. We labeled 104 training images sampled uniformly from our dataset using the nine categories stated in Sec. \ref{sec:semantic}.

We used a scaled version ($512\times256$) of the image for training in order to handle large VRAM requirement. We present our used parameters in Table \ref{tab:segnetparameters} for reference.

\begin{table}[h]
	\begin{center}
		\begin{tabular}{|l|c|}
			\hline
			Parameter & Value \\
			\hline\hline
			learning rate & 0.001\\
			gamma & 1.0 \\
			momentum & 0.9 \\
			weight decay & 0.0005 \\
			iteration & 10000\\
			batch size & 4 \\
			dropout ratio & 0.5 \\
			output batch size & 8 \\
			\hline
		\end{tabular}
	\end{center}
	\caption{Parameter setting for training on a Bayesian SegNet Basic Model.}
	\label{tab:segnetparameters}
\end{table}

\begin{figure}
	\begin{center}
		\includegraphics[width=1.0\linewidth]{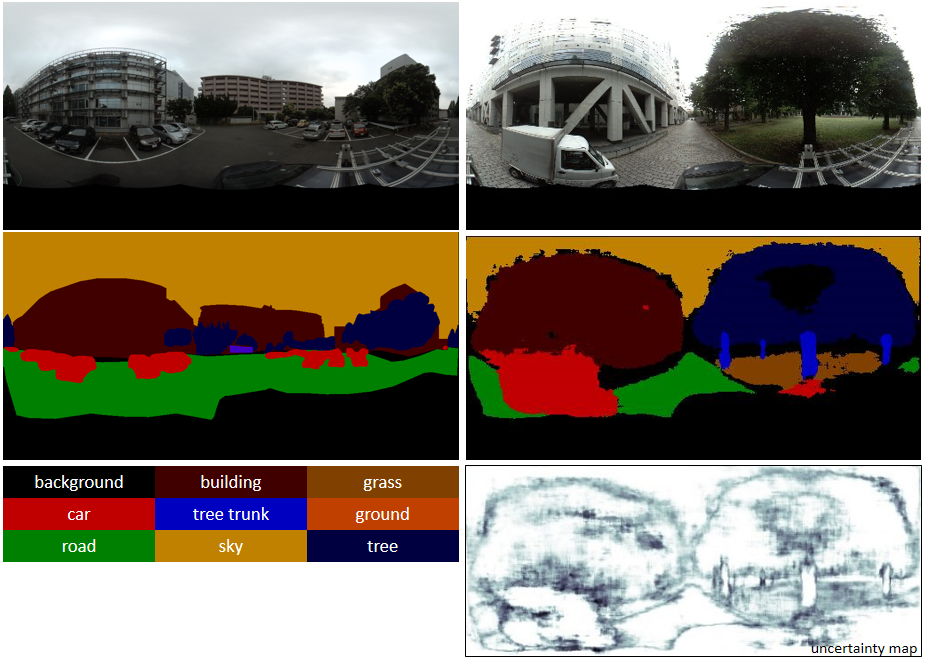}
	\end{center}
	\caption{Input frames and labels for training (left column) and result of segmentation (right column) with colors corresponding to different classes (bottom left) and the uncertainty of classification (bottom right)(white = high certainty).}
	\label{fig:semanticsegmentation}
\end{figure}

We achieved a classification accuracy of $88.56\%$ with training time of $181 s$. Using the learned model, we feed each frame on the network and achived classification time of 310 ms. The output of each processing is the semantic segmentation of the original frame into nine classes including the uncertainty of the classification. Figure \ref{fig:semanticsegmentation} shows sample frame for the manually labeled images and the result of the semantic segmentation with uncertainty frame.

\subsection{Optical Flow Estimation}
We implemented a version of the TV-L1 \cite{tvl1} optical flow method on a Intel Core i7-4930K and GTX1080Ti GPU using CUDA. TV-L1 optical flow estimation achieves real-time results with reasonable accuracy. Moreover, the Total Variation (TV) regularization used in this method can handle discontinuities in motion estimation that usually happens along object boundaries. This greatly benefits our foreground probability map estimation because we want the boundaries to be as accurate as possible.

Following the notations in the paper (See \cite{tvl1} for details), we set the following parameters fitted for our dataset: $\lambda=100, \theta=0.3, \tau=0.25, iteration=115$. We also use a pyramid scaling of $0.5$ and $6$ pyramid levels. We are able to achieve a frame rate of $15 fps$ on a $1024\times512$ image which is suitable for our application.

\section{Results and Experiments}
\label{sec:results}
\subsection{Comparison with existing methods}
We compare the results of our method with simple alpha blending and bi-stable transparency blending \cite{fukiage} methods. For all three methods, we use the same depth map to solve the foreground probability map. For the alpha blending method we solve the color of the pixel as:

\begin{equation}
\label{eq:alphablending}
RGB = Real_{RGB} \times P_f + Cg_{RGB} \times (1-P_f)
\end{equation}

For \cite{fukiage}, we set the $V_f$ and $V_b$ as fixed based on the region class (see Table \ref{tab:transparency}), and solve the visiblity as in Equation \ref{eq:visibility}.

\begin{table}
	\begin{center}
		\begin{tabular}{|l|c|c|c|c|c|c|}
			\hline
			Category & $V_f$ \cite{fukiage} & $V_b$ \cite{fukiage} & $V_{f1}$ & $V_{f2}$ & $V_{b1}$ & $V_{b2}$ \\
			\hline\hline
			Background & 10.0 & 10.0 & 10.0 & 10.0 & 10.0 & 10.0 \\
			Simple & 0.0005 & 5.0 & 0.0005 & 0.001 & 5.0 & 4.0 \\
			Complex & 1.5 & 4.0 & 1.5 & 1.0 & 4.0 & 2.5 \\
			\hline
		\end{tabular}
	\end{center}
	\caption{Visibility parameters setting for \cite{fukiage} and our method.}
	\label{tab:transparency}
\end{table}

\begin{figure*}
	\begin{center}
		\includegraphics[width=1.0\linewidth]{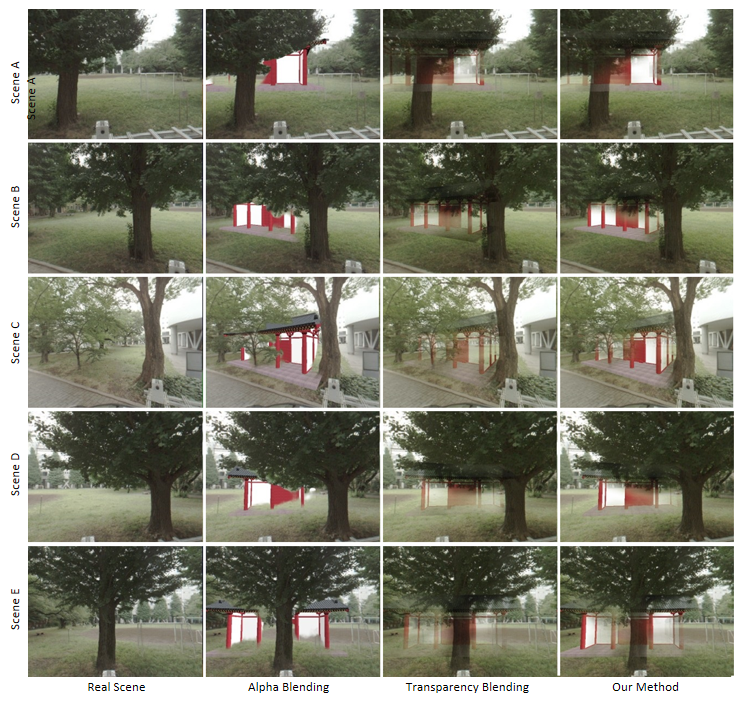}
	\end{center}
	\caption{Comparison of rendering results from alpha blending, transparency blending, and our method.}
	\label{fig:comparison}
\end{figure*}

We show the comparison of the output from the three methods in Figure \ref{fig:comparison}. The first column correspond to the frame seen by the HMD. The second, third and fourth column are the output of the alpha blending, transparency blending and our method. In all cases, the alpha blending method achieves the highest visibility value. However, it is apparent along the more complex contours of tree leaves that the alpha blending fails. The method results in an insufficient segmentation of the foreground region.

In contrast, the transparency blending achieves more visually pleasing blending along the complex contours. However, the visibility of the virtual object suffers when the background is smooth. This results in the virtual object being almost invisible.

Our method achieves the best tradeoff between visibility and accurate segmentation. Along the regions of the complex contours of the foreground, our method outperforms the simple alpha blending. When the background is flat, our method outperforms the transparency blending. 

\begin{figure*}
	\begin{center}
		\includegraphics[width=0.8\linewidth]{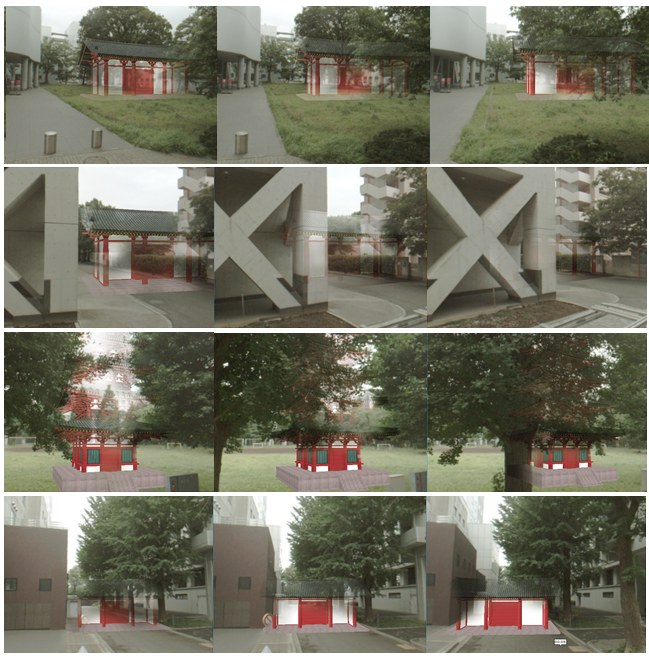}
	\end{center}
	\caption{Results of our occlusion handling method on different scenes. Top to bottom: different scenarios. Left to right: succeeding frames.}
	\label{fig:moreresults}
\end{figure*}

We present more results from our method in Figure \ref{fig:moreresults}. Our method depends on the correctness of the semantic segmentation. For example in the third row, the sky was incorrectly classified therefore appearing more visible. Moreover, due to the lack of texture, the foreground probability map was also inaccurate. This issue can be rectified by using more training data for the semantic segmentation. Furthermore, the training can be overfitted to the environment where the MR system will be deployed. 

\subsection{User Evaluation}
Using the same settings for the three methods as in the previous section, we conducted an experiment with users (6 male and female, ages 23-48). Five scenes (see Figure \ref{fig:comparison}) of 10-second video each were randomly shown to the users. We performed a pairwise comparison (total of 6 combinations) among the three methods. We showed one sequence first and then another and asked the users to compare the two sequence based on three categories: 1)Visibility of virtual object (Is it easy to see the virtual object?), 2)Realistic occlusion of the virtual object (Does the virtual object appear to be realistically occluded?) and 3) Realistic appearance of the rendered scene (Does the scene look realistic?). Each of the sequence is graded from -3 to +3 (+3 if the second video has maximum preferrence, -3 if the first video has maximum preference). We also randomly show each video pairs in reverse order, resulting in 30 pairs of evaluation dataset.

\begin{figure*}
	\begin{center}
		\includegraphics[width=0.9\linewidth]{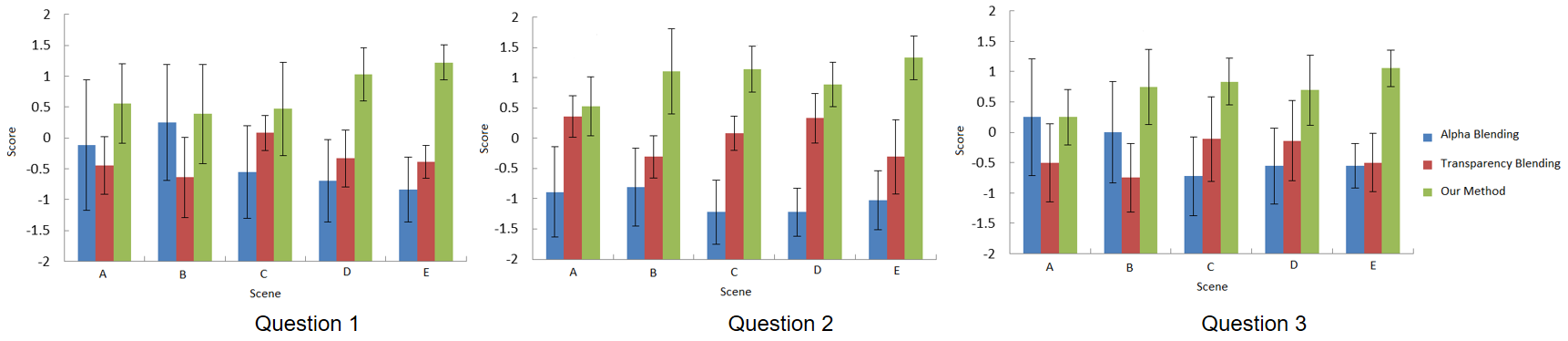}
	\end{center}
	\caption{Comparison of preference scores for Question 1 (visibility of virtual object), Question 2 (realistic occlusion of the virtual object) and Question 3 (realistic appearance of the rendered scene).}
	\label{fig:usereval}
\end{figure*}

Based on the evaluation, we plot the total preference scores for each scene and questions in Figures \ref{fig:usereval}. In all the tests, our method achieved highest preferential scores compared to the other two methods.

\section{Conclusion and Future Work}
\label{sec:conclusion}
In this work, we demonstrated how to use visibility-based blending method in handling occlusion problem in mixed reality. We incorporated depth estimation from optical flow to solve the foreground probability map together with semantic classsification using convolutional neural network. Our results shows that compared to existing alpha blending and transparency blending based techniques, our method achieves better visibility in flat background areas and better occlusion handling along complex foreground objects.

However, limitations in the semantic segmentation only allows us to achieve sub-real-time processing. In the future, a faster implementation of semantic segmentation that can perform in real-time is desired. Furthermore, a more robust camera pose estimation that handles real-time and outdoor applications is also desired.

\nocite{*}

%

\bibliographystyle{ACM-Reference-Format}
\bibliography{template} 


\begin{thebibliography}{00}


\ifx \showCODEN    \undefined \def \showCODEN     #1{\unskip}     \fi
\ifx \showDOI      \undefined \def \showDOI       #1{#1}\fi
\ifx \showISBNx    \undefined \def \showISBNx     #1{\unskip}     \fi
\ifx \showISBNxiii \undefined \def \showISBNxiii  #1{\unskip}     \fi
\ifx \showISSN     \undefined \def \showISSN      #1{\unskip}     \fi
\ifx \showLCCN     \undefined \def \showLCCN      #1{\unskip}     \fi
\ifx \shownote     \undefined \def \shownote      #1{#1}          \fi
\ifx \showarticletitle \undefined \def \showarticletitle #1{#1}   \fi
\ifx \showURL      \undefined \def \showURL       {\relax}        \fi
\providecommand\bibfield[2]{#2}
\providecommand\bibinfo[2]{#2}
\providecommand\natexlab[1]{#1}
\providecommand\showeprint[2][]{arXiv:#2}

\bibitem[\protect\citeauthoryear{andK. Takemoto, Sato, Yamamoto, and
  Tamura}{andK. Takemoto et~al\mbox{.}}{2002}]%
        {uchiyama}
\bibfield{author}{\bibinfo{person}{S.~Uchiyama andK. Takemoto},
  \bibinfo{person}{K. Sato}, \bibinfo{person}{H. Yamamoto}, {and}
  \bibinfo{person}{H. Tamura}.} \bibinfo{year}{2002}\natexlab{}.
\newblock \showarticletitle{MR Platform: A basic body on which mixed reality
  applications are built}. In \bibinfo{booktitle}{{\em ISMAR}}.
\newblock


\bibitem[\protect\citeauthoryear{Badrinarayanan, Handa, and
  Cipolla}{Badrinarayanan et~al\mbox{.}}{2015}]%
        {segnet}
\bibfield{author}{\bibinfo{person}{Vijay Badrinarayanan},
  \bibinfo{person}{Ankur Handa}, {and} \bibinfo{person}{Roberto Cipolla}.}
  \bibinfo{year}{2015}\natexlab{}.
\newblock \showarticletitle{SegNet: A Deep convolutional encoder-decoder
  architecture for robust semantic pixel-wise labelling}. In
  \bibinfo{booktitle}{{\em CVPR}}.
\newblock


\bibitem[\protect\citeauthoryear{Criminisi, Cross, Blake, and
  Kolmogorov}{Criminisi et~al\mbox{.}}{2006}]%
        {criminisi}
\bibfield{author}{\bibinfo{person}{A. Criminisi}, \bibinfo{person}{G. Cross},
  \bibinfo{person}{A. Blake}, {and} \bibinfo{person}{V. Kolmogorov}.}
  \bibinfo{year}{2006}\natexlab{}.
\newblock \showarticletitle{Bilayer segmentation of live video}. In
  \bibinfo{booktitle}{{\em Computer Vision and Pattern Recognition}}.
  \bibinfo{pages}{53--60}.
\newblock


\bibitem[\protect\citeauthoryear{Fukiage, Oishi, and Ikeuchi}{Fukiage
  et~al\mbox{.}}{2012}]%
        {fukiage}
\bibfield{author}{\bibinfo{person}{Taiki Fukiage}, \bibinfo{person}{Takeshi
  Oishi}, {and} \bibinfo{person}{Katsushi Ikeuchi}.}
  \bibinfo{year}{2012}\natexlab{}.
\newblock \showarticletitle{Reduction of contradictory partial occlusion in
  mixed reality by using characteristics of transparency perception}. In
  \bibinfo{booktitle}{{\em ISMAR}}. \bibinfo{pages}{129--139}.
\newblock


\bibitem[\protect\citeauthoryear{Fukiage, Oishi, and Ikeuchi}{Fukiage
  et~al\mbox{.}}{2014}]%
        {fukiage2}
\bibfield{author}{\bibinfo{person}{Taiki Fukiage}, \bibinfo{person}{Takeshi
  Oishi}, {and} \bibinfo{person}{Katsushi Ikeuchi}.}
  \bibinfo{year}{2014}\natexlab{}.
\newblock \showarticletitle{Visibility-based blending for real-time
  applications}. In \bibinfo{booktitle}{{\em ISMAR}}. \bibinfo{pages}{63--72}.
\newblock


\bibitem[\protect\citeauthoryear{Kakuta, L.B.Vinh, Kawakami, T.Oishi, and
  K.Ikeuchi}{Kakuta et~al\mbox{.}}{2008}]%
        {kakuta}
\bibfield{author}{\bibinfo{person}{T. Kakuta}, \bibinfo{person}{L.B.Vinh},
  \bibinfo{person}{R. Kawakami}, \bibinfo{person}{T.Oishi}, {and}
  \bibinfo{person}{K.Ikeuchi}.} \bibinfo{year}{2008}\natexlab{}.
\newblock \showarticletitle{Detection of moving objects and cast shadows using
  spherical vision camera for outdoor mixed reality}. In
  \bibinfo{booktitle}{{\em VRST}}. \bibinfo{pages}{219--222}.
\newblock


\bibitem[\protect\citeauthoryear{Kanbara, Okuma, Takemura, and Yokoya}{Kanbara
  et~al\mbox{.}}{1999}]%
        {depth2}
\bibfield{author}{\bibinfo{person}{Masayuki Kanbara}, \bibinfo{person}{Takashi
  Okuma}, \bibinfo{person}{Haruo Takemura}, {and} \bibinfo{person}{Naokazu
  Yokoya}.} \bibinfo{year}{1999}\natexlab{}.
\newblock \showarticletitle{Real-time composition of stereo images for video
  see-through augmented reality}. In \bibinfo{booktitle}{{\em International
  Conference on Multimedia Computing and Systems}}. \bibinfo{pages}{213--219}.
\newblock


\bibitem[\protect\citeauthoryear{Kim, jun Yang, and Sohn}{Kim
  et~al\mbox{.}}{2005}]%
        {depth3}
\bibfield{author}{\bibinfo{person}{Hansung Kim}, \bibinfo{person}{Seung jun
  Yang}, {and} \bibinfo{person}{Kwanghoon Sohn}.}
  \bibinfo{year}{2005}\natexlab{}.
\newblock \showarticletitle{3D reconstruction of stereo images for interaction
  between real and virtual objects}. In \bibinfo{booktitle}{{\em Signal
  Processing and Image Communication}}. \bibinfo{pages}{61--75}.
\newblock


\bibitem[\protect\citeauthoryear{Kim, Jung, Lee, and Lee}{Kim
  et~al\mbox{.}}{2008}]%
        {depth4}
\bibfield{author}{\bibinfo{person}{Tae~Hoon Kim}, \bibinfo{person}{Hoyub Jung},
  \bibinfo{person}{Kyoung~Mu Lee}, {and} \bibinfo{person}{Sang~Uk Lee}.}
  \bibinfo{year}{2008}\natexlab{}.
\newblock \showarticletitle{Segmentation based foreground object disparity
  estimation using Zcam and multiple-view stereo}. In \bibinfo{booktitle}{{\em
  International Conference on Intelligent Information Hiding and Multimedia
  Signal Processing}}. \bibinfo{pages}{1251--1254}.
\newblock


\bibitem[\protect\citeauthoryear{LabelMe}{LabelMe}{[n. d.]}]%
        {labelme}
\bibfield{author}{\bibinfo{person}{LabelMe}.} \bibinfo{year}{[n.
  d.]}\natexlab{}.
\newblock \bibinfo{title}{LabelMe. The Open Annotation Tool}.
\newblock \bibinfo{howpublished}{http://labelme.csail.mit.edu}.
  (\bibinfo{year}{[n. d.]}).
\newblock


\bibitem[\protect\citeauthoryear{L.B.Vinh, Kakuta, Kawakami, Oishi, and
  Ikeuchi}{L.B.Vinh et~al\mbox{.}}{2010}]%
        {vinh}
\bibfield{author}{\bibinfo{person}{L.B.Vinh}, \bibinfo{person}{Tetsuya Kakuta},
  \bibinfo{person}{Rei Kawakami}, \bibinfo{person}{Takeshi Oishi}, {and}
  \bibinfo{person}{Katsushi Ikeuchi}.} \bibinfo{year}{2010}\natexlab{}.
\newblock \showarticletitle{Foreground and shadow occlusion handling for
  outdoor augmented reality}. In \bibinfo{booktitle}{{\em ISMAR}}.
  \bibinfo{pages}{13--16}.
\newblock


\bibitem[\protect\citeauthoryear{Li, Sun, Tang, and Shum}{Li
  et~al\mbox{.}}{2006}]%
        {lazysnap}
\bibfield{author}{\bibinfo{person}{Yin Li}, \bibinfo{person}{Jian Sun},
  \bibinfo{person}{Chi-Keung Tang}, {and} \bibinfo{person}{Heung-Yeung Shum}.}
  \bibinfo{year}{2006}\natexlab{}.
\newblock \showarticletitle{Lazy Snapping}. In \bibinfo{booktitle}{{\em ACM
  Transaction on Graphics}}, Vol.~\bibinfo{volume}{23}.
  \bibinfo{pages}{22--32}.
\newblock


\bibitem[\protect\citeauthoryear{Liu and He}{Liu and He}{2015}]%
        {liu}
\bibfield{author}{\bibinfo{person}{B. Liu} {and} \bibinfo{person}{X. He}.}
  \bibinfo{year}{2015}\natexlab{}.
\newblock \showarticletitle{Multiclass Semantic video segmentation with
  object-level active inference}. In \bibinfo{booktitle}{{\em CVPR}}.
\newblock


\bibitem[\protect\citeauthoryear{Long, Shelhamer, and Darrell}{Long
  et~al\mbox{.}}{2015}]%
        {long}
\bibfield{author}{\bibinfo{person}{J. Long}, \bibinfo{person}{E. Shelhamer},
  {and} \bibinfo{person}{T. Darrell}.} \bibinfo{year}{2015}\natexlab{}.
\newblock \showarticletitle{Fully convolutional networks for semantic
  segmentation}. In \bibinfo{booktitle}{{\em CVPR}}.
\newblock


\bibitem[\protect\citeauthoryear{Papagiannakis and Schertenleib}{Papagiannakis
  and Schertenleib}{2005}]%
        {papagiannakis}
\bibfield{author}{\bibinfo{person}{George Papagiannakis} {and}
  \bibinfo{person}{Sebastien Schertenleib}.} \bibinfo{year}{2005}\natexlab{}.
\newblock \showarticletitle{Mixing virutal and real scenes in the site of
  ancient Poppeii}. In \bibinfo{booktitle}{{\em Journal of Computer Animation
  and Virtual Worlds}}. \bibinfo{pages}{11--24}.
\newblock


\bibitem[\protect\citeauthoryear{Research}{Research}{[n. d.]}]%
        {ladybug}
\bibfield{author}{\bibinfo{person}{Point~Grey Research}.} \bibinfo{year}{[n.
  d.]}\natexlab{}.
\newblock \bibinfo{title}{Lady Bug Technical Reference}.
\newblock
  \bibinfo{howpublished}{https://www.ptgrey.com/ladybug5-360-degree-usb3-spherical-camera-systems}.
    (\bibinfo{year}{[n. d.]}).
\newblock


\bibitem[\protect\citeauthoryear{Rother, Kolmogorov, and Blake}{Rother
  et~al\mbox{.}}{2004}]%
        {grabcut}
\bibfield{author}{\bibinfo{person}{Carsten Rother}, \bibinfo{person}{Vladimir
  Kolmogorov}, {and} \bibinfo{person}{Andrew Blake}.}
  \bibinfo{year}{2004}\natexlab{}.
\newblock \showarticletitle{Grab-cut - Interactive foreground extraction using
  iterated graph cuts}. In \bibinfo{booktitle}{{\em ACM, SIGGRAPH '04}}.
  \bibinfo{pages}{309--314}.
\newblock


\bibitem[\protect\citeauthoryear{SegNet}{SegNet}{[n. d.]}]%
        {segnetwebsite}
\bibfield{author}{\bibinfo{person}{SegNet}.} \bibinfo{year}{[n.
  d.]}\natexlab{}.
\newblock \bibinfo{title}{SegNet: A Deep convolutional encoder-decoder
  architecture for robust semantic pixel-wise labelling}.
\newblock \bibinfo{howpublished}{http://mi.eng.cam.ac.uk/projects/segnet}.
  (\bibinfo{year}{[n. d.]}).
\newblock


\bibitem[\protect\citeauthoryear{Sharma, Tuzel, and Jacobs}{Sharma
  et~al\mbox{.}}{2015}]%
        {sharma}
\bibfield{author}{\bibinfo{person}{A. Sharma}, \bibinfo{person}{O. Tuzel},
  {and} \bibinfo{person}{D. Jacobs}.} \bibinfo{year}{2015}\natexlab{}.
\newblock \showarticletitle{Deep Hierarchical parsing for semantic
  segmentation}. In \bibinfo{booktitle}{{\em CVPR}}.
\newblock


\bibitem[\protect\citeauthoryear{Sun, Zhang, Tang, and H.Y.Shum}{Sun
  et~al\mbox{.}}{2006}]%
        {backgroundcut}
\bibfield{author}{\bibinfo{person}{J. Sun}, \bibinfo{person}{W. Zhang},
  \bibinfo{person}{X. Tang}, {and} \bibinfo{person}{H.Y.Shum}.}
  \bibinfo{year}{2006}\natexlab{}.
\newblock \showarticletitle{Background Cut}. In \bibinfo{booktitle}{{\em
  ECCV}}. \bibinfo{pages}{628--641}.
\newblock


\bibitem[\protect\citeauthoryear{S.W.Hasinoff, Kang, and
  R.Szeliski}{S.W.Hasinoff et~al\mbox{.}}{2006}]%
        {boundarymatting}
\bibfield{author}{\bibinfo{person}{S.W.Hasinoff}, \bibinfo{person}{S.B. Kang},
  {and} \bibinfo{person}{R.Szeliski}.} \bibinfo{year}{2006}\natexlab{}.
\newblock \showarticletitle{Boundary Matting for View Synthesis}. In
  \bibinfo{booktitle}{{\em Computer Vision and Image Understanding}}.
  \bibinfo{pages}{53--60}.
\newblock


\bibitem[\protect\citeauthoryear{Varas, Alfaro, and Marques}{Varas
  et~al\mbox{.}}{2015}]%
        {varas}
\bibfield{author}{\bibinfo{person}{D. Varas}, \bibinfo{person}{M. Alfaro},
  {and} \bibinfo{person}{F. Marques}.} \bibinfo{year}{2015}\natexlab{}.
\newblock \showarticletitle{Mutiresolution hierarchy co-clustering for semantic
  segmentation in sequences with small variations}. In \bibinfo{booktitle}{{\em
  ICCV}}.
\newblock


\bibitem[\protect\citeauthoryear{Zach, Pock, and Bishof}{Zach
  et~al\mbox{.}}{2007}]%
        {tvl1}
\bibfield{author}{\bibinfo{person}{C. Zach}, \bibinfo{person}{T. Pock}, {and}
  \bibinfo{person}{H. Bishof}.} \bibinfo{year}{2007}\natexlab{}.
\newblock \showarticletitle{A duality based approach for real-time TV-L1
  optical flow}. In \bibinfo{booktitle}{{\em Lecture Notes in Computer Science:
  Pattern Recognition}}.
\newblock


\bibitem[\protect\citeauthoryear{Zhu, Wang, Yang, and Davis}{Zhu
  et~al\mbox{.}}{2008}]%
        {depth1}
\bibfield{author}{\bibinfo{person}{Jiejie Zhu}, \bibinfo{person}{Liang Wang},
  \bibinfo{person}{Ruigang Yang}, {and} \bibinfo{person}{James Davis}.}
  \bibinfo{year}{2008}\natexlab{}.
\newblock \showarticletitle{Fusion of time-of-flight depth and stereo for high
  accuracy depth maps}. In \bibinfo{booktitle}{{\em CVPR}}.
  \bibinfo{pages}{23--28}.
\newblock


\bibitem[\protect\citeauthoryear{Zollmann and Reitmayr}{Zollmann and
  Reitmayr}{2012}]%
        {zollmann}
\bibfield{author}{\bibinfo{person}{S. Zollmann} {and} \bibinfo{person}{G.
  Reitmayr}.} \bibinfo{year}{2012}\natexlab{}.
\newblock \showarticletitle{Dense Depth Maps from Sparse Models and Image
  Coherence for Augmented Reality}. In \bibinfo{booktitle}{{\em VRST '12}}.
\newblock


\end{thebibliography}

\end{document}